\documentclass{article}
\pdfoutput=1
\usepackage[final, nonatbib]{nips_2016}
\usepackage[numbers]{natbib}

\usepackage[utf8]{inputenc} 
\usepackage[T1]{fontenc}    
\usepackage{hyperref}       
\usepackage{url}            
\usepackage{booktabs}       
\usepackage{amsfonts}       
\usepackage{nicefrac}       
\usepackage{microtype}      
\usepackage{amsmath}

\DeclareMathOperator{\Cor}{Cor}
\DeclareMathOperator*{\argmax}{arg\,max}

\title{Canonical Correlation Analysis for Analyzing Sequences of Medical Billing Codes}

\author{
  Corinne~L.~Jones\\
  Department of Statistics\\
 University of Washington\\
 Seattle, WA 98195 \\
 \texttt{cjones6@uw.edu} \\
 \And
     Sham~M.~Kakade\\
  Departments of Statistics and \\ Computer Science \& Engineering \\
 University of Washington\\
 Seattle, WA 98195 \\
 \texttt{sham@cs.washington.edu}
 \And
 Lucas~W.~Thornblade \\
 Department of Surgery\\
 University of Washington\\
 Seattle, WA 98105 \\
  \texttt{lucaswt@uw.edu} \\
 \And
 David~R.~Flum \\
  Department of Surgery\\
 University of Washington\\
 Seattle, WA 98105 \\
  \texttt{daveflum@uw.edu} \\
 \And 
  Abraham~D.~Flaxman\\
  Institute for Health Metrics and Evaluation\\
  University of Washington\\
  Seattle, WA 98121 \\
  \texttt{abie@uw.edu} \\
}

\begin{document}

\maketitle

\begin{abstract}
We propose using canonical correlation analysis (CCA) to generate features from sequences of medical billing codes. Applying this novel use of CCA to a database of medical billing codes for patients with diverticulitis, we first demonstrate that the CCA embeddings capture meaningful relationships among the codes. We then generate features from these embeddings and establish their usefulness in predicting future elective surgery for diverticulitis, an important marker in efforts for reducing costs in healthcare.
\end{abstract}

\section{Introduction}
Datasets with sequences of medical billing codes assigned to patients over time have the potential to provide a lot of insight into many areas, including questions in epidemiology, comparative effectiveness, and discovery. The primary obstacle to using such data sets in machine learning lies in feature generation. In particular, inputting these codes into algorithms as one-hot vectors misses the inherent relationships that exist between many of the codes. In this paper we demonstrate that the application of canonical correlation analysis (CCA) \citep{hotelling1936} can generate features that capture relationships between the codes. To our knowledge, this is the first application of CCA to medical claims data.

We then use the CCA features in the downstream task of predicting elective surgery for diverticulitis. Diverticulitis, a disease characterized by inflamed pouches in the colon called diverticula, affects roughly 88 in 100,000 Americans each year \citep{wheat2016}. The burden of healthcare associated with diverticulitis is significant, accounting for more than 300,000 hospitalizations and \$2.6 billion in annual healthcare expenditures \citep{etzioni2009}. While diverticulitis is often treated with antibiotics, rates of elective surgery for this condition have increased three-fold in recent years \cite{simianu2016}. In an effort to reduce potentially low-value surgery, surgeons seek to characterize patients who may be at highest risk for undergoing future elective surgery. Using healthcare claims data, we aim in this paper to predict future elective surgery in diverticulitis patients by inputting features generated with CCA into machine learning algorithms. If doctors can predict who would be more likely to undergo elective surgery, they may try harder to dissuade these patients from doing so.

We briefly review CCA in Section~\ref{sec:background} and describe how we can apply it to medical billing codes in Section~\ref{cca_mine}. Following this, in Section~\ref{sec:experiments} we discuss the data set we use and present results  comparing our CCA features to one-hot encodings of codes (unigram features) and pairs of codes (bigram features) on several machine learning algorithms when predicting elective surgery for diverticulitis.

\section{\label{sec:background} Background}
Many methods exist to meaningfully project features into a lower-dimensional space. One such method is canonical correlation analysis \citep{hotelling1936}, which has been recently discussed in the natural language processing literature \citep{stratos2014},  \citep{stratos2015}. Following \cite{stratos2015}, the idea behind canonical correlation analysis (CCA) is, given mean-zero random vectors $X\in \mathbb{R}^n$ and $Y\in\mathbb{R}^{n'}$, to maximize the correlation between $X$ and $Y$ in a lower dimensional space. For our purposes it is easiest to think of $X$ and $Y$ as two descriptions of a medical billing code. To achieve this, we solve the problem 
\begin{align*}
(a_i, b_i) = &\argmax_{a \in \mathbb{R}^n, b\in \mathbb{R}^{n'}} \Cor(a^TX, b^TY) \\
&\text{s.t.} \ \Cor(a^TX, a_j^TX) = 0 \ \ \forall \ j<i\\
& \hspace{.52cm} \Cor(b^TY, b_j^TY) = 0\ \ \ \  \forall \ j<i
\end{align*}
for $a_1,\dots,a_m\in \mathbb{R}^n$ and $b_1,\dots,b_m\in \mathbb{R}^{n'}$. 
Combining the projection vectors into matrices $A=[a_1|a_2|\cdots|a_m]$ and $B=[b_1|b_2|\cdots|b_m]$, we then can write the projections of $X$ and $Y$ as $\tilde X = A^TX \in \mathbb{R}^m$ and $\tilde Y = B^TY\in \mathbb{R}^m$. Hotelling \cite{hotelling1936} showed that $A$ and $B$ equal the $m$ left and right singular vectors, respectively, corresponding to the $m$ largest singular values of 
$\Omega \equiv E[XX^T]^{-1/2}E[XY^T]E[YY^T]^{-1/2}.$

In relating CCA to the problem of generating word embeddings, the authors in \cite{stratos2015} define $X$ and $Y$ to be one-hot vectors of words and their ``contexts'', respectively. In their paper, ``context'' consists of a word within a distance $\rho$ of the current word.
The authors then note that for a large number of samples the elements of the matrix $\Omega$ may be estimated as 
$\hat \Omega_{w,c} = \frac{\hat P(X_w=1, Y_c=1)}{\sqrt{\hat P(X_w=1)\hat P(Y_c=1)}}$. Here $X_w$ represents word $w$ while $Y_c$ represents context $c$.

One other consideration when performing CCA with word embeddings is that some words may occur very few times in the data set. This causes the estimated matrices $\hat E[XX^T]$ and $\hat E[YY^T]$ to be ill-conditioned. Several methods exist to deal with this, including  removing infrequently observed words and using regularization. \cite{stratos2015} takes the first approach in its experiments and replaces all words occurring less than 100 times in the corpus with a single new symbol. 

In this paper, we generate embeddings of medical billing codes using canonical correlation analysis. Other approaches to creating embeddings found in the natural language processing literature include the neural-based methods of \cite{mikolov2013} and \cite{ling2015}.  However, unlike with the neural methods, an elegant justification exists for the CCA embeddings, as noted in \cite{stratos2014} and \cite{stratos2015}. On downstream tasks \cite{stratos2015} shows that CCA performs competitively with the neural methods and outperforms other spectral approaches. That paper also shows that adding CCA features leads to a drastic increase in performance over simply using one-hot vectors for words. 

\section{\label{cca_mine} Methodology}
In applying CCA to sequences of medical billing codes, we propose treating the codes similarly to words in sentences as in the natural language processing literature, but with several modifications. Naively performing CCA on medical codes where we define the random variable $X$ to be the diagnosis, drug, or procedure code for a patient and we define the random variable $Y$ to be a code occurring at most $\rho$ places before or after it for that patient makes little sense. First, there could be multiple codes on a given day for a patient, and so then the order of the codes on that day is likely meaningless. Second, and more importantly, if the code we see today is the first code that person has had in five years, that code five years ago is relatively unlikely to be related to today's code. On the other hand, a code used yesterday is much more likely to be related to today's code. Hence, it would make more sense to define context by proximity of dates assigned to codes rather than by the order in which they were assigned.

With that in mind, we include every pair of codes for an individual in estimating the matrix $E[XY^T]$. Still using one-hot encodings to represent each code $x$ and context $y$, we weight each pair $(x,y)$ by the ``distance'' from $x$ to $y$, where ``distance'' refers to how long it was between the time that $x$ was observed and the time that $y$ was observed. Additionally, we ensure that each person in the data set contributes equal weight to the estimated matrix $\hat E[XY^T]$. In particular, we set the weights for person $p$ and observed code $x$ and context $y$ to be 
$\tilde w_p(x,y) = \frac{w_p(x,y)}{\sum_{x,y}w_p(x,y)},$
where
 $w_p(x,y) = \frac{1}{1+\# \text{weeks between code } x \text{ and context } y}$.
For example, if two codes occur on the same day for a given person, then, ignoring the division by the total weight for person $p$, that pair of codes gets weight 1 added to their corresponding element of the matrix $\hat E[XY^T]$. If they were a week apart, they would get weight 0.5. Dividing by the total weight for the person ensures that every person contributes equally to the matrix $\hat E[XY^T]$. To obtain the estimated matrix $\hat E[XX^T]$, we simply diagonalize the row sums of $\hat E[XY^T]$, and do the same for $\hat E[YY^T]$ with the column sums.

\section{Experiments}\label{sec:experiments}
In our experiments we first perform CCA on medical claims data using the modifications discussed above. We then transform the results into input for algorithms that predict which patients will undergo future elective surgery for diverticulitis. The first subsection describes the data used in this analysis, while the second subsection discuss the results and their implications.

\subsection{Data}
In our experiments, we use medical claims data for the years 2007-2014 from a Truven Health MarketScan\textsuperscript{\textregistered} Research Database. This database includes person-level healthcare utilization data for employees and dependents covered by employer-sponsored private health insurance for more than 30 million Americans annually. We selected patients who developed diverticulitis during that time period and  all associated diagnostic, drug and procedure codes with their corresponding dates. Although the database also contains some demographic information, we found that including this data in the prediction task made little difference in the results. Because of this, and since our focus lies in generating features from the billing codes, we report the predictions without those covariates.

The entire data set contains 265 million medical billing codes (76,000 unique codes) across 770,000 patients. For predicting elective surgery, we restrict ourselves to patients who were in the data set for at least two continuous years before and after their initial diverticulitis diagnosis, who did not have surgery for cancer, who did not have emergency surgery for diverticulitis, and who did not have surgery during the 52 weeks following their initial diagnosis. We predict those patients going on to elective surgery between 52 and 104 weeks following their initial diverticulitis diagnosis based on their medical codes up through 48 weeks after diagnosis. In total, 82,231 patients satisfy the above criteria, 947 of whom received elective surgery between one and two years after diagnosis. In performing CCA, we use all of the codes from patients not included in the prediction task, along with all of the codes up through 48 weeks after diagnosis for those who were included in the prediction task. This amounts to 240 million codes.

\subsection{Results}
In our experiments we first perform CCA as described in Section~\ref{cca_mine} and project the codes to a 25-dimensional space. We regularize the estimated matrices $\hat E[XX^T]$ and $\hat E[YY^T]$ by adding  $\lambda_x=0.01\times \max(\hat E[XX^T])$ and $\lambda_y=0.01\times \max(\hat E[YY^T])$ to their diagonals, respectively. 

To show how well CCA captures the relationships between codes, we display in Table~\ref{tab:proj_reg1} the nearest three neighbors (based on Euclidean distance) of several codes: icd9-56211 (diverticulitis of colon) and two arbitrarily selected codes unrelated to diverticulitis: icd9-19882 (malignant neoplasm of temporal lobe) and ndc-00093028093 (bupropion HCl).
It is evident from the table that CCA discovers real relationships, as, e.g., diverticulitis patients have abdominal pain. However, it is not always perfect, as seen in the results for bupropion HCl: bupropion HCl, an antidepressant, seems like it should be less closely related to lisinopril, a drug for high blood pressure and heart failure. Nevertheless, it makes sense that it would be related to clonazepam (a drug for seizures and panic disorders) and a fracture of the base of the skull. 

\begin{table}[t!]
  \caption{\label{tab:proj_reg1} Nearest neighbors to several codes after CCA when using regularization parameters $\lambda_x=0.01\times \max(\hat E[XX^T])$ and $\lambda_y=0.01\times \max(\hat E[YY^T])$ and projecting to 25 dimensions}
\begin{center}
\begin{tabular}{ll}
Code (description) & Nearest neighbors \\ \hline
icd9-56211 (Diverticulitis of colon) & icd9-78900 (Abdominal pain, unspecified site)\\ 
&icd9-78904 (Abdominal pain, lower left quadrant)\\
& cpt-72193 (CT pelvis with contrast)\\ 
icd9-19882 (Malignant neoplasm of temporal lobe)& icd9-1910 (Malignant neoplasm of frontal lobe)  \\
&icd9-1710(Malignant neoplasm of soft tissue of head) \\
& icd9-1808 (Malignant neoplasm of cervix NEC)\\ 
ndc-00093028093 (Bupropion HCl) & ndc-24658024530 (Lisinopril) \\
& ndc-60505006803 (Clonazepam)\\
& icd9-80141 (Fracture of base of skull)
\end{tabular}
\end{center}
\end{table}

Given the CCA projections, we converted the codes for a patient into features by averaging the projections corresponding to their codes before their initial diverticulitis diagnosis and averaging the projections for the codes after their initial diagnosis. After standardizing the features, we input them into the scikit-learn\cite{scikit-learn} implementations of gradient boosting, $L_2$-regularized logistic regression, and random forest (with 1000 trees). To select the hyperparameters, we perform 3-fold cross validation over the values 1, 3, 5,\dots, 11 for the maximum depth of the classifiers in gradient boosting and over the default set of regularization parameters in penalized logistic regression. As a comparison to the CCA features, we also try using only unigram features, unigram and bigram features, and  CCA and unigram and bigram features. We limit the unigram features to the codes occurring at least 10 times in the data set and bigram features to pairs of adjacent codes occurring at least 10 times. 

\begin{table}[t!]
  \caption{\label{tab:pred} Mean (standard deviation) of AUC scores from 10-fold cross-validation comparing CCA features to unigram and bigram features in the prediction of elective surgery for diverticulitis.}
\begin{center}
\begin{tabular}{lcccc}
Algorithm & Unigram & Unigram \& Bigram & CCA & All\\ \hline
Gradient boosting & 0.75 (0.03) & 0.76 (0.03) &0.74 (0.04)& 0.76 (0.03) \\
Penalized logistic regression & 0.58 (0.02) & 0.56 (0.02) & 0.74 (0.03)& 0.58 (0.03)\\
Random forest& 0.72 (0.02) & 0.74 (0.02) & 0.72 (0.03) & 0.76 (0.03)
\end{tabular}
\end{center}
\end{table}

Table~\ref{tab:pred} presents the average and standard deviation of the AUC scores from 10-fold cross-validation. Gradient boosting, on average, performs as well as or better than the other classifiers for each feature set. In addition, gradient boosting with the 50 CCA features comes within one standard deviation of the AUC for gradient boosting with unigram and bigram features. However, the CCA features drastically outperform the unigram and bigram features in penalized logistic regression. Training times on all of the algorithms are much faster when using the 50 CCA features compared to the 20,000 unigram features and 240,000 unigram \& bigram features.

In light of the recent surge in neural network approaches, we also considered a multi-layer perceptron with 100 hidden units and regularization parameter 1.0. Preliminary results produced average AUCs of 0.71, 0.70, 0.73, and 0.74 for unigram, unigram \& bigram, CCA, and all features, respectively. Improving on the parameter choices for each feature set might be a fruitful direction for future study.

\section{Conclusion}
In this paper, we presented a novel application of canonical correlation analysis, showing that CCA can be used to generate features from sequences of medical billing codes. Applying this to a data set containing medical billing codes for diverticulitis patients, we demonstrated that CCA can capture meaningful relationships between the codes. Moreover, our results suggest that using just 50 features derived from averaged CCA projections can significantly improve predictions from penalized logistic regression. These features also achieve AUC scores within one standard deviation of the best feature set with gradient boosting and random forest, but with the benefit of faster training times. Future work may include comparing CCA to neural network embeddings in this context.


\section*{Acknowledgments}
Research reported in this publication was supported by the National Institute Of Diabetes And Digestive And Kidney Diseases under Award Number R01DK103915. The content is solely the responsibility of the authors and does not necessarily represent the official views of the National Institutes of Health.


\bibliographystyle{plainnat}
\bibliography{references}

\end{document}